\documentclass[cameraready]{Interspeech}

\usepackage{graphicx}
\usepackage{subcaption}
\usepackage{multirow}
\usepackage{booktabs}
\usepackage{algorithm}
\usepackage{algorithmic}
\usepackage{xspace}

\usepackage{amsmath,amsfonts,bm}

\def\eqref#1{equation~\ref{#1}}

\def\1{\bm{1}}

\DeclareMathAlphabet{\mathsfit}{\encodingdefault}{\sfdefault}{m}{sl}
\SetMathAlphabet{\mathsfit}{bold}{\encodingdefault}{\sfdefault}{bx}{n}

\newcommand{\algo}{WAND\xspace}

\title{WAND: Windowed Attention and Knowledge Distillation\\ 
for Efficient Autoregressive Text-to-Speech Models}

\author[affiliation={1}, orcid=0009-0009-2253-0874]{Hanna}{Lee}
\author[affiliation={1}, orcid=0009-0005-7599-1720]{Tan Dat}{Nguyen}
\author[affiliation={2}, orcid=0009-0004-4117-8371]{Jaehoon}{Kang}
\author[affiliation={2}, orcid=0000-0002-0123-3100]{Kyuhong}{Shim}

\address{
    $^1$ Korea Advanced Institute of Science and Technology, Republic of Korea \\
    $^2$ Sungkyunkwan University, Republic of Korea
}

\email{
    hanna.lee@kaist.ac.kr,
    tandat.kaist@kaist.ac.kr,
    morateng@skku.edu,
    khshim@skku.edu
}

\keywords{text-to-speech, speech synthesis, efficient TTS, global-local attention, \algo}

\usepackage{comment}

\begin{document}

\maketitle

\begin{abstract}
Recent decoder-only autoregressive text-to-speech (AR-TTS) models produce high-fidelity speech, but their memory and compute costs scale quadratically with sequence length due to full self-attention.
In this paper, we propose WAND, Windowed Attention and Knowledge Distillation, a framework that adapts pretrained AR-TTS models to operate with constant computational and memory complexity.
WAND separates the attention mechanism into two: persistent global attention over conditioning tokens and local sliding-window attention over generated tokens.
To stabilize fine-tuning, we employ a curriculum learning strategy that progressively tightens the attention window.
We further utilize knowledge distillation from a full-attention teacher to recover high-fidelity synthesis quality with high data efficiency.
Evaluated on three modern AR-TTS models, WAND preserves the original quality while achieving up to 66.2\% KV cache memory reduction and length-invariant, near-constant per-step latency.
\end{abstract}

\section{Introduction}

\begin{figure*}[t!]
    \centering
    \includegraphics[width=\linewidth]{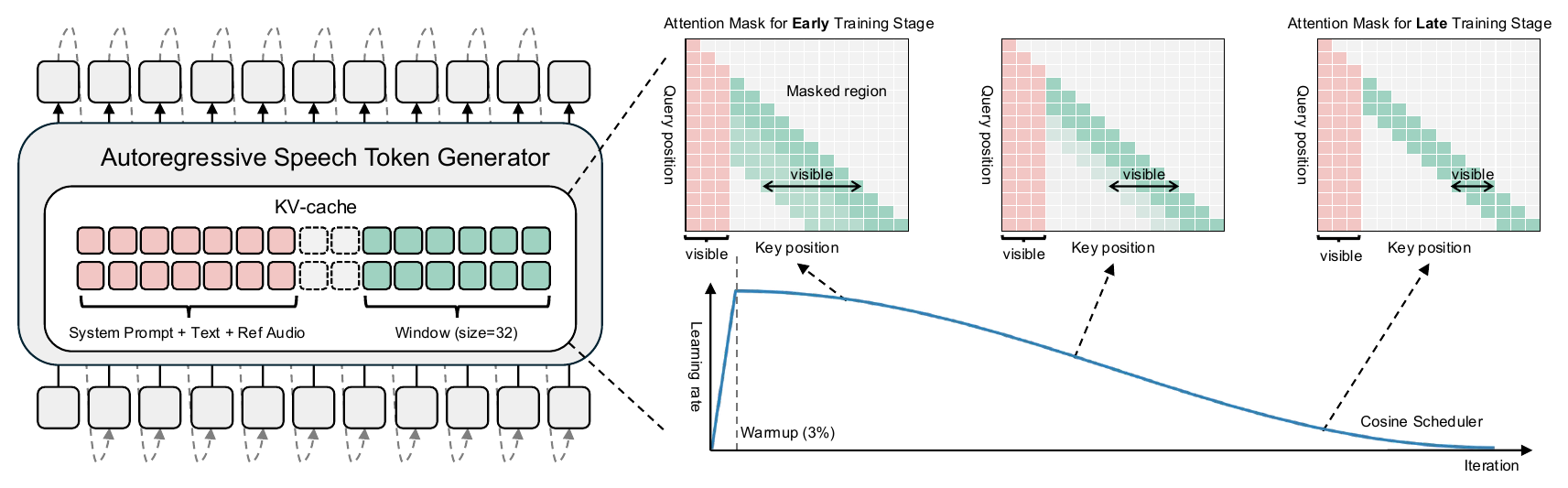}
    \caption{Overview of the \algo framework. Conditioning tokens (system prompt, text, reference audio) retain global attention access, while generated acoustic tokens are restricted to a fixed-size sliding window of size $W$.}
    \label{fig:method}
\end{figure*}

Text-to-Speech (TTS) models leveraging Transformer-based large language model (LLM) backbones have demonstrated the capability to generate near-human-quality speech from short reference audio~\cite{valle,sparktts,cosyvoice2,deng2025indextts}. 
By utilizing neural audio codecs~\cite{defossez2022high,zeghidour2021soundstream}, these models treat speech as discrete token sequences, enabling the direct adaptation of text-based LLMs for speech synthesis.
Compared to conventional seq2seq TTS models such as Tacotron~\cite{wang2017tacotron} and VITS~\cite{kim2021conditional}, LLM-based TTS systems exhibit superior zero-shot speaker generalization, greater robustness to diverse linguistic inputs, and enhanced naturalness.
However, the computational complexity of Transformer architectures scales quadratically with sequence length~\cite{vaswani2017attention}, leading to substantial memory overhead and high inference latency.
This challenge hinders the deployment of LLM-based TTS models in stringent real-time applications~\cite{park2024long}.\looseness=-1

To circumvent the scaling limitations of the Transformer, recent research has explored several avenues for reducing the computational intensity and memory requirements of LLM architectures. 
One prominent direction focuses on reducing the model depth through layer-wise pruning \cite{gromov2024unreasonable,shortgpt}, an approach recently applied to speech synthesis to minimize parameter counts~\cite{spade}. 
However, depth reduction fails to address the fundamental nature of the self-attention mechanism, as the remaining layers still incur quadratic computational costs and identical per-layer memory overhead. 
On the other hand, architectural shifts toward linear-time mechanisms, such as Gated Linear Attention~\cite{yang2024gated,linaspeech} and Mamba-based architectures~\cite{mamba2,DBLP:conf/icassp/JiangLFHM25,nguyen2026mamtra}, offer significantly higher throughput. 
However, these paradigms necessitate training from scratch and often exhibit a performance gap, where the resulting speech quality and prosodic naturalness remain inferior to those achieved by established AR-TTS models.
Other approaches accelerate inference by optimizing the decoding process, such as through speculative decoding~\cite{leviathan2023fast, nguyen2025accelerating, li2025fast} or parallel decoding~\cite{santilli2023accelerating, kou2024cllms}.
While this improves speed by generating multiple tokens per step, it still relies on self-attention and leaves the underlying memory scaling issues unresolved.
A more common and simpler alternative is the use of Key-Value (KV) caching~\cite{pope2022efficiently,kwon2023efficient}, which is widely adopted in state-of-the-art models~\cite{cosyvoice2,sparktts}. 
While KV caching achieves a near-linear inference cost similar to linear attention while preserving the original model's performance, it introduces a critical bottleneck:
the cumulative memory footprint continues to expand with every generated token~\cite{kim2024infinipot,liu2025reattention}. 
This restricts the synthesis of long-form utterances and imposes severe hardware constraints, a challenge that current LLM-based TTS systems have yet to resolve effectively.\looseness=-1

In this work, we hypothesize that \textit{AR-TTS models may not require full-sequence attention to maintain high-fidelity synthesis}. 
Specifically, we posit that the conditioning prompt tokens, comprising target text, reference audio, and auxiliary task or emotion tags, provide sufficient semantic and acoustic context for the generation. 
In contrast, generated speech tokens primarily serve to maintain local temporal consistency; thus, the model only requires a \textit{localized view} of these tokens to track the decoding position relative to the monotonic speech trajectory.
This is consistent with the ``attention sink'' phenomenon in text LLMs~\cite{streamingllm}, where most of the attention mass is concentrated on several prefix tokens and a local window.
Despite this, current AR-TTS architectures attend to all tokens indiscriminately, resulting in a linearly expanding KV cache that inherently limits long-sequence scalability.\looseness=-1

Motivated by this view, we propose WAND, \textbf{W}indowed \textbf{A}ttention and K\textbf{n}owledge \textbf{D}istillation, to transform the scaling of computational and memory costs from linear to constant.
Our framework bifurcates the attention mechanism into two distinct components:
\textit{Global Attention}, which maintains persistent access to the conditioning tokens (capturing 48--65\% of attention mass), and \textit{Local Sliding-Window Attention}, which restricts the receptive field for generated speech tokens to a fixed-size window. 
Unlike paradigms that necessitate training from scratch, \algo leverages pretrained weights, substantially reducing training cost and enabling seamless integration into existing AR-TTS pipelines.
By employing knowledge distillation from a full-attention teacher, \algo recovers high-fidelity synthesis quality using only 53.8 hours of speech data~\cite{hinton2015distilling}.~\looseness=-1

We evaluate \algo across three different models:
CosyVoice~2~\cite{cosyvoice2}, IndexTTS~1.5~\cite{deng2025indextts}, and SparkTTS~\cite{sparktts}.
For 10-second generation, \algo achieves up to a 66.2\% reduction in KV cache size and significantly decreases cumulative computational costs, reducing total GFLOPs by as much as 46.9\%.
Notably, \algo achieves effective adaptation using only 53.8 hours of English data in a single epoch and generalizes to Mandarin without training data with a Character Error Rate (CER) degradation within 0.1\% absolute. 
Our contributions are threefold:
(1)~an attention restriction method for LLM-based TTS ensuring constant memory and computational overhead without architectural modifications; (2)~a data-efficient adaptation strategy via knowledge distillation that generalizes across languages; and (3)~a cross-architecture validation achieving a bounded KV cache and constant per-step latency with negligible quality loss.
We make our source code and demo publicly available.~\footnote{\url{https://onemeee.github.io/wand-tts/}}~\looseness=-1

\begin{table*}[!t]
\centering
\caption{Comparison of baselines and \algo models on \texttt{test-en}. 
Inference efficiency metrics (KV cache and GFLOPs) correspond to the accumulation required to generate 10 seconds of audio, with the KV cache reported in fp32 mode. 
All \algo models are fine-tuned solely on 53.8 hours of English data. 
NMOS is reported with a confidence interval of $95\%$.
}
\label{tab:main-results}
\vspace{-0.2cm}
\renewcommand{\arraystretch}{1.15}
\setlength{\tabcolsep}{5pt}
\resizebox{\linewidth}{!}{
\begin{tabular}{ll|cc|rc|cc|lcll}
\toprule
\textbf{Model} & \textbf{Attn. Mechanism} & \textbf{Cache} & \textbf{Compute} & \textbf{KV Cache $\downarrow$} & \textbf{Reduction $\uparrow$} & \textbf{GFLOPs $\downarrow$} & \textbf{Speedup $\uparrow$} & \textbf{UTMOS $\uparrow$} & \textbf{NMOS $\uparrow$} & \textbf{SSIM $\uparrow$} & \textbf{WER $\downarrow$} \\
\midrule
Groundtruth & $-$ & $-$ & $-$ & $-$ & $-$ & $-$ & $-$ & $3.53$ & $3.79\pm0.18$ & $100$ & $1.44$ \\
\midrule
\multirow{2}{*}{CosyVoice2-0.5B} & GQA ($W=\infty$) & $\mathcal{O}(L)$ & $\mathcal{O}(L)$ & $10.48$ MB & $-$ & $11.55$ & $-$ & $4.18$ & $4.01\pm0.17$ & $\mathbf{94.6}$ & $1.94$ \\
                                 & WAND ($W=32$)         & $\mathcal{O}(1)$ & $\mathcal{O}(1)$ & $\mathbf{5.25}$ MB & $\mathbf{49.9\%}$ & $\mathbf{7.44}$ & $\mathbf{1.55\times}$ & $\mathbf{4.21_{+0.03}}$ & $4.02\pm0.16$ & $94.5_{-0.1}$ & $\mathbf{1.72_{-0.22}}$ \\
\midrule
\multirow{2}{*}{IndexTTS 1.5}    & MHA ($W=\infty$) & $\mathcal{O}(L)$ & $\mathcal{O}(L)$ & $38.44$~MB & $-$ & $6.18$ & $-$ & $3.97$ & $4.13\pm0.14$ & $\mathbf{94.4}$ & $0.98$ \\
                                 & WAND ($W=32$)         & $\mathcal{O}(1)$ & $\mathcal{O}(1)$ & $\mathbf{13.01}$~MB & $\mathbf{66.2\%}$ & $\mathbf{3.28}$ & $\mathbf{1.89\times}$ & $\mathbf{3.98_{+0.01}}$ & $4.13\pm0.13$ & $\mathbf{94.4_{+0.0}}$ & $\mathbf{0.91_{-0.07}}$ \\
\midrule
\multirow{2}{*}{SparkTTS-0.5B}   & GQA ($W=\infty$) & $\mathcal{O}(L)$ & $\mathcal{O}(L)$ & $18.09$~MB & $-$ & $48.12$ & $-$ & $\mathbf{3.93}$ & $3.94\pm0.16$ & $92.6$ & $3.27$ \\
                                 & WAND ($W=64$)         & $\mathcal{O}(1)$ & $\mathcal{O}(1)$ & $\mathbf{7.15}$~MB & $\mathbf{60.5\%}$ & $\mathbf{31.74}$ & $\mathbf{1.51\times}$ & $\mathbf{3.93_{+0.00}}$ & $3.97\pm0.16$ & $\mathbf{92.8_{+0.2}}$ & $\mathbf{3.11_{-0.16}}$ \\
\bottomrule
\end{tabular}}
\end{table*}
\section{Method}

\subsection{Attention Restriction with Sliding Window Decoding}

We posit that AR-TTS relies on two distinct types of information:
\textit{global context}, which determines the invariant characteristics of the generation, and \textit{local context}, which governs fine-grained acoustic transitions~\cite{yoon2023pruning,spade,kang2026unlocking}.
By decoupling these, we can restrict the model's temporal attention to a local window without losing global consistency or speaker identity.
Without loss of generality, we consider TTS conditioned on four fundamental inputs: a system prompt $s$ that encodes global constraints, target text $x$, a reference audio prompt $a^{\mathrm{pr}}$ specifying speaker style, and the generated acoustic tokens $y_{1:T}$.
These inputs are concatenated into a conditioning prompt:
\begin{equation}
p_\theta(y_{1:T}\mid s,x,a^{\mathrm{pr}}) = \prod_{t=1}^{T} p_\theta\left(y_t \mid y_{<t}, s, x, a^{\mathrm{pr}}\right).
\end{equation}
As shown in Figure~\ref{fig:method}, we divide the attention mechanism into two distinct components.
\textit{Global Attention} (red blocks in Figure~\ref{fig:method}) provides fixed conditioning on $s$, $x$, and $a^{\mathrm{pr}}$.
Conversely, \textit{Local Sliding-Window Attention} (green blocks in Figure~\ref{fig:method}) provides dynamic conditioning on the recent acoustic history $y_{<t}$.
Because acoustic signals are locally coherent and monotonic, the influence of distant past tokens $y_{1:t-W-1}$ diminishes once global conditions are fixed.
By restricting local attention to a window $W$, we not only maintain efficiency but also mitigate the propagation of sampling artifacts from the distant past:
\begin{equation}p_\theta(y_t \mid y_{<t}, s, x, a^{\mathrm{pr}}) \approx p_\theta\left(y_t \mid y_{t-W:t-1}, s, x, a^{\mathrm{pr}}\right).
\end{equation}

By partitioning the KV cache into a fixed global component and a rolling window of size $W$ for acoustic tokens, we achieve a constant $\mathcal{O}(1)$ inference memory footprint relative to the total sequence length $T$.
At the inference stage, this constant-length windowed attention enables arbitrarily long audio generation with bounded memory usage.\looseness=-1

\subsection{Knowledge Distillation for Attention Adaptation}

While our hypothesis suggests that global context maintains core identity, some performance degradation is inevitable when a model trained on full attention is suddenly restricted to a local window.
Our initial experiments show that it results in a slight decrease in content consistency.
To mitigate this observed performance gap, we employ a knowledge distillation framework that supervises the restricted student model using two complementary objectives~\cite{hinton2015distilling,inaguma2021alignment,yoon2025heuristic}.
First, a cross-entropy loss ($\mathcal{L}_{\mathrm{CE}}$) anchors the student to the ground-truth acoustic tokens, ensuring fundamental alignment with the target speech.
Second, a Skew Kullback–Leibler divergence~\cite{ko2024distillm} loss ($\mathcal{L}_{\mathrm{KL}}$) encourages the token probability distribution of the student to mimic the probability of a full-attention teacher model, both conditioned on the same historical context.
This loss helps the student remain consistent when the long-range context is removed after attention adaptation.
The final training objective is a weighted combination of both components:
$
\mathcal{L} = \mathcal{L}_{\mathrm{CE}} + \lambda\,\mathcal{L}_{\mathrm{KL}}
$.

\subsection{Curriculum Scheduling for Window Reduction}
\label{sec:curr_sched}

To stabilize fine-tuning, we introduce a curriculum that progressively reduces the window size from $W_{\text{start}}$ to the target size $W$.
Following a cosine schedule $\alpha(t)$ over $T_c$ steps:
\begin{equation}\alpha(t) = \frac{1}{2}\left(1 - \cos\left(\pi \min\left(\frac{t}{T_c}, 1\right)\right)\right),\end{equation}
the effective window size at step $t$ is defined as $W(t) = W_{\text{start}} - \alpha(t)\left(W_{\text{start}} - W\right)$.
To further ease adaptation, we apply a temperature-controlled soft mask to the attention logits $A$.
Rather than immediate hard truncation, we define $A' = A - \tau(t)M$, where $M$ is the binary mask for positions outside the current window.
The masking strength $\tau(t)$ follows a log-scale interpolation:
\begin{equation}
\tau(t) = \exp\Big(\log \tau_{\text{start}} + \alpha(t)\left(\log \tau_{\text{end}} - \log \tau_{\text{start}}\right)\Big).
\end{equation}
As depicted in Figure~\ref{fig:method}, an initially small $\tau(t)$ permits partial attention to masked positions, preserving gradient flow during early optimization.
As $\tau(t)$ increases, the model gradually adapts to strict inference-time constraints.
\section{Experiments}
\textbf{Datasets \& Fine-Tuning Configurations.}\,\,\,\,We conduct fine-tuning experiments on the \texttt{train-clean-100} subset of LibriTTS~\cite{libritts}, which contains approximately 53.8 hours of transcribed speech.
Notably, we intentionally use only this modest subset rather than the full 585.80-hour corpus, demonstrating that \algo requires minimal data for effective adaptation.
All models are fine-tuned using the AdamW optimizer with a cosine learning rate schedule and a peak learning rate of $1 \times 10^{-5}$.
We fine-tune each baseline with WAND for 1 epoch on a single NVIDIA A100 MIG partition with 20\,GB of memory.\looseness=-1

\vspace{0.15cm}\noindent\textbf{Baselines.}\,\,\,\,We evaluate \algo on three pretrained AR-TTS models spanning different architectures and codec configurations:
\textbf{CosyVoice~2-0.5B}~\cite{cosyvoice2}:
A 0.5B-parameter model built on the Qwen2.5-0.5B backbone, using an FSQ-based single-codebook codec with a token rate of 25\,Hz;
\textbf{IndexTTS 1.5}~\cite{deng2025indextts}:
A GPT-style AR-TTS model employing a single-codebook VQ codec with a token rate of 25\,Hz;
and \textbf{SparkTTS-0.5B}~\cite{sparktts}: 
A 0.5B-parameter model built on the Qwen2.5-0.5B~\cite{qwen2024qwen2} backbone, utilizing BiCodec with a semantic token rate of 50\,Hz.
This diversity in backbone architectures, codec designs (FSQ, VQ, BiCodec), and token rates (25\,Hz vs.\ 50\,Hz) provides a rigorous testbed for evaluating the generality of WAND.\looseness=-1

\begin{table}
\centering
\caption{Attention distribution of vanilla (full-attention) AR-TTS models during decoding, averaged over 5 long utterances.
}
\vspace{-0.2cm}
\label{tab:attention-analysis}
\renewcommand{\arraystretch}{1.1}
\resizebox{\linewidth}{!}{
\begin{tabular}{l|cc|c|c}
\toprule
\multirow{2}{*}{\textbf{Model}} 
& \multicolumn{2}{c|}{\textbf{Attention (\%)}} 
& \textbf{Local-$W$}
& \multirow{2}{*}{\textbf{Cov.}} \\
& \textbf{Prompt} & \textbf{Generated} & \textbf{/Gen. (\%)} & \\
\midrule
CosyVoice2 ($W=32$)   & 58.5 & 41.5 & 70.2 & 87.6 \\
IndexTTS 1.5 ($W=32$) & 64.6 & 35.4 & 57.1 & 84.8 \\
SparkTTS ($W=64$)      & 47.9 & 52.1 & 82.8 & 91.0 \\
\bottomrule
\end{tabular}
}
\end{table}

\vspace{0.15cm}\noindent\textbf{Evaluations.}\,\,\,\,We evaluate WAND using the Seed-TTS-eval~\cite{seedtts} benchmark, performing tests on both \texttt{test-en} and \texttt{test-zh} to assess generation quality on both English and Mandarin.\footnote{\url{https://github.com/BytedanceSpeech/seed-tts-eval}} 
For objective metrics, we calculate Word Error Rate (WER) using Whisper-large-v3~\cite{radford2023robust} and Character Error Rate (CER) using Paraformer-zh~\cite{funasr} as ASR systems to evaluate content accuracy for English and Mandarin, respectively.
We also measure the speaker similarity (SSIM) as the cosine similarity between speaker embeddings extracted from the reference prompt and the synthesized speech using WavLM-based~\cite{chen2022wavlm} x-vectors, while overall speech quality is assessed via the UTMOS metric~\cite{saeki2022utmos}.
Subjectively, we conduct Naturalness Mean Opinion Score (NMOS) tests on a 5-point scale involving 10 raters, with 50 randomly generated audio samples from each model to provide a measure of perceived naturalness.\looseness=-1

\section{Analysis}

\subsection{Target-Language Speech Quality}

As demonstrated in Table~\ref{tab:main-results}, WAND maintains high-fidelity synthesis across all evaluated architectures on the Seed-TTS \texttt{test-en} benchmark.
Notably, the absolute word error rate (WER) either remains within 0.2\% of the baseline or exhibits slight improvement, as observed in CosyVoice~2, where the WER decreased from 1.94\% to 1.72\%.
We attribute this performance stability---and occasional improvement--- to the inherent regularization effect of the sliding window.
By isolating the decoding context, WAND effectively mitigates the propagation of distant sampling artifacts and hallucinations.
Consequently, this limits the model's dependency on its own prior outputs, mitigating the compounding errors typical of autoregressive generation, and forces the student model to rely more robustly on the invariant global conditioning.

\subsection{Inference Efficiency}

\begin{figure}[t]
    \centering
    \includegraphics[width=\linewidth]{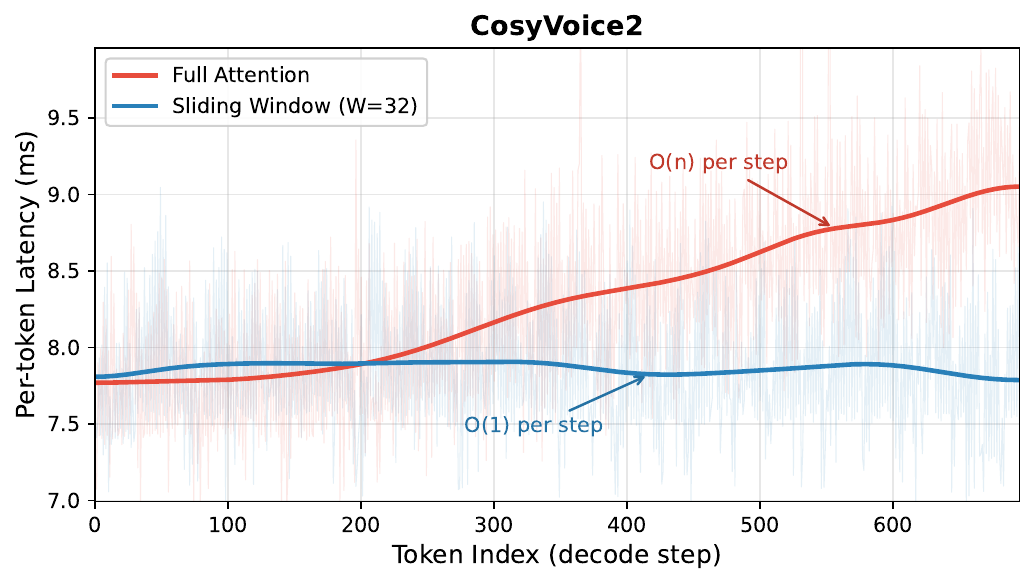}
    \caption{Per-step decoding latency regarding the number of generated tokens. Full attention latency grows linearly with sequence length, while sliding-window attention maintains constant latency regardless of output length.}
    \label{fig:latency}
    \vspace{-1ex}
\end{figure}

Beyond accuracy, WAND delivers substantial efficiency gains.
For 10-second generation, memory usage is reduced by up to $66.2$\%, with IndexTTS~1.5 KV cache ($fp32$) decreasing from $38.44$MB to $13.01$MB.
Computational cost is similarly reduced: CosyVoice~2 GFLOPs drop from $11.55$ to $7.44$, and IndexTTS from $6.18$ to $3.28$, yielding speedups of $1.51$$\times$–$1.89$$\times$.
CosyVoice~2 and SparkTTS employ Grouped Query Attention~\cite{ainslie2023gqa}.
FLOPs are computed using the full number of query heads, as KV heads are broadcast during attention. 
SparkTTS exhibits higher absolute FLOPs due to its higher token rate, resulting in double decoding steps compared to other baselines.
The theoretical memory reduction translates directly to inference efficiency (Fig.~\ref{fig:latency}).
While full attention latency increases from $7.8$ms to over $9.0$ms with sequence length, \algo maintains near-constant latency due to bounded attention.
Combined with bounded KV cache growth, this enables long-form synthesis with constant memory and constant per-step computation.\looseness=-1

\subsection{Data Efficiency and Cross-Lingual Preservation}

Our framework achieves effective adaptation with only 53.8 hours of data and a single epoch of fine-tuning.
Crucially, despite this English-only fine-tuning stage, \algo preserves the ability to generate high-quality Mandarin.
As shown in Table~\ref{tab:results-zh}, the performance on Chinese speech remains robust, with character error rate (CER) degradation limited to within 0.1\% absolute for primary systems like CosyVoice~2 and IndexTTS~1.5.
This cross-lingual resilience confirms that the learned attention restriction captures a universal structural property of speech (i.e., temporal locality) rather than language-specific patterns.
Furthermore, \algo generalizes across three architecturally diverse systems---differing in backbone (Qwen2.5-based vs.\ GPT-style), codec design (FSQ, VQ, BiCodec), and token rate (25\,Hz, 50\,Hz)---without any architecture-specific modifications.
This broad applicability confirms that the attention restriction is a universal optimization applicable to the AR-TTS paradigm, enabling consistent efficiency gains regardless of the underlying codec or model configuration.

\subsection{Attention Pattern Analysis}

\begin{table}[t]
\centering
\caption{Comparison of baselines and their fine-tuned version with \algo on \texttt{test-zh}. All \algo models are fine-tuned exclusively on English data.\looseness=-1}
\label{tab:results-zh}
\vspace{-0.2cm}
\renewcommand{\arraystretch}{1.1}
\resizebox{\linewidth}{!}{
\begin{tabular}{ll|ccc}
\toprule
\textbf{Model} & \textbf{Attn. Mechanism} & \textbf{UTMOS} $\uparrow$ & \textbf{SSIM} $\uparrow$ & \textbf{CER} $\downarrow$ \\
\midrule
\multirow{2}*{CosyVoice~2-0.5B} & Baseline ($W=\infty$) & 3.49 & 95.3 & 1.59 \\
 & \algo ($W=32$) & 3.50 & 95.3 & 1.53 \\
\midrule
\multirow{2}*{IndexTTS 1.5} & Baseline ($W=\infty$) & 3.23 & 95.6 & 0.82 \\
 & \algo ($W=32$) & 3.25 & 95.6 & 0.91 \\
\midrule
\multirow{2}*{SparkTTS-0.5B} & Baseline ($W=\infty$) & 3.28 & 94.2 & 2.14 \\
 & \algo ($W=64$) & 3.26 & 94.4 & 2.35 \\
\bottomrule
\end{tabular}}
\vspace{-1ex}
\end{table}

To validate the design of our attention restriction, we analyze the attention patterns of vanilla full-attention models (Table~\ref{tab:attention-analysis}).
Across all three architectures, 48--65\% of attention targets the conditioning prefix, confirming that these tokens serve as persistent anchors throughout generation and must retain global access.
Within the decode region, 57--83\% of attention concentrates on the most recent $W$ tokens.
Combining these components, the prefix and a local window jointly account for 85--91\% of total attention, validating that restricting decode-region attention to a fixed window discards only a small fraction of the model's attention mass.
Notably, SparkTTS allocates a larger share to the decode region (52\%) and requires a proportionally wider window ($W$\!=\!64) to achieve comparable coverage, consistent with its higher codec token rate (50\,Hz).

\subsection{Ablation Studies}

\begin{table}[t]
\caption{Ablation on distillation loss components. WER (\%) on Seed-TTS-eval \texttt{test-en}.}
\label{tab:loss-ablation}
\vspace{-0.2cm}
\renewcommand{\arraystretch}{1.1}
\resizebox{\linewidth}{!}{
\begin{tabular}{l|c|c|c}
\toprule
\multirow{2}{*}{\textbf{Loss}}
& \textbf{CosyVoice~2}
& \textbf{IndexTTS 1.5}
& \textbf{SparkTTS} \\
& $W$=32 & $W$=32 & $W$=64 \\
\midrule
No distillation (SW only)  & 3.40 & 1.11 & 3.81 \\
$\mathcal{L}_{\mathrm{CE}}$ only          & 2.02 & 1.16 & 4.92 \\
$\mathcal{L}_{\mathrm{KL}}$ only          & 2.37 & 1.05 & 3.83 \\
$\mathcal{L}_{\mathrm{CE}} + \mathcal{L}_{\mathrm{KL}}$  & \textbf{1.72} & \textbf{0.91} & \textbf{3.11} \\
\bottomrule
\end{tabular}
}
\end{table}

\textbf{Distillation Loss Components.}\,\,\,\,Table~\ref{tab:loss-ablation} isolates the contribution of each loss component.
Applying the sliding window without distillation achieves a reasonable WER of 1.11\% for IndexTTS~1.5, but substantially degrades performance for CosyVoice~2 (3.40\%) and SparkTTS (3.81\%). While adding either $\mathcal{L}_{\mathrm{CE}}$ or $\mathcal{L}_{\mathrm{KL}}$ alone recovers a portion of the performance gap for some models, only their combination consistently yields the best results across all three models.
The result confirms that cross-entropy and KL divergence losses provide complementary supervision during attention adaptation.\looseness=-1

\vspace{0.15cm}\noindent\textbf{Curriculum Learning.}\,\,\,\,Table~\ref{tab:curriculum-ablation} compares direct training with the target window size against our curriculum strategy.
The curriculum strategy begins with a window size of 128 and progressively narrows to the target $W$ using temperature-scaled masking to gradually restrict attention over distant tokens. 
This approach consistently outperforms direct training, reducing WER from 1.86\% to 1.72\% for CosyVoice~2 and from 1.03\% to 0.91\% for IndexTTS~1.5.\looseness=-1

\begin{table}[t]
\caption{Ablation on curriculum learning strategy. WER (\%) on Seed-TTS-eval \texttt{test-en}. Curriculum progressively reduces window size from 128 to $W$ with temperature-scaled masking.}
\vspace{-0.2cm}
\renewcommand{\arraystretch}{1.1}
\setlength{\tabcolsep}{10pt}
\label{tab:curriculum-ablation}
\resizebox{\linewidth}{!}{
\begin{tabular}{l|c|c}
\toprule
\multirow{2}{*}{\textbf{Training Strategy}}
& \textbf{CosyVoice~2}
& \textbf{IndexTTS 1.5} \\
& $W$=32 & $W$=32 \\
\midrule
Direct ($W=32$ from start)      & 1.86 & 1.03 \\
Curriculum ($W=128 \to 32$)     & 1.72 & 0.91 \\
\bottomrule
\end{tabular}}
\vspace{-1ex}
\end{table}
\section{Conclusion}

In conclusion, \textbf{WAND} successfully transforms the computational and memory scaling of AR-TTS from linear to constant by bifurcating the attention mechanism into persistent \textit{Global Attention} and \textit{Local Sliding-Window Attention}.
Our evaluations across CosyVoice~2, IndexTTS~1.5, and SparkTTS demonstrate that bounded KV caching enables constant per-step latency with negligible quality loss.
Specifically, the framework achieved up to a 66.2\% reduction in KV cache size and a significant decrease in GFLOPs, while maintaining robust cross-lingual generalization even with limited data.
By overcoming the memory scaling limits inherent in Transformers, WAND enables long-form TTS with constant memory and per-step computation, paving the way for continuous, infinitely long audio generation without hardware constraints or performance degradation.
\clearpage

\section{Acknowledgments}
This work was supported by Institute of Information \& communications Technology Planning \& Evaluation (IITP) grant funded by the Korea government (MSIT) (RS-2019-II190421, Artificial Intelligence Graduate School Program (Sungkyunkwan University)).

\section{Generative AI Use Disclosure}
Generative AI tools (Claude, Anthropic) were used to assist with grammar correction and polishing of the manuscript. 
All scientific content, experimental design, and analysis were conducted solely by the authors. 
All authors take full responsibility for the content of this paper.

\bibliographystyle{IEEEtran}
\bibliography{shortstrings,mybib}

@inproceedings{vaswani2017attention,
  title={Attention is All You Need},
  author={Vaswani, Ashish and Shazeer, Noam and Parmar, Niki and Uszkoreit, Jakob and Jones, Llion and Gomez, Aidan N and Kaiser, {\L}ukasz and Polosukhin, Illia},
  booktitle={Advances in Neural Information Processing Systems},
  volume={30},
  year={2017}
}

@article{valle,
  title={Neural Codec Language Models are Zero-Shot Text to Speech Synthesizers},
  author={Wang, Chengyi and Chen, Sanyuan and Wu, Yu and Zhang, Ziqiang and Zhou, Long and Liu, Shujie and Chen, Zhuo and Liu, Yanqing and Wang, Huaming and Li, Jinyu and He, Lei and Zhao, Sheng and Wei, Furu},
  journal={arXiv preprint arXiv:2301.02111},
  year={2023}
}

@article{cosyvoice2,
  title={CosyVoice 2: Scalable Streaming Speech Synthesis with Large Language Models},
  author={Du, Zhihao and others},
  journal={arXiv preprint arXiv:2412.10117},
  year={2024}
}

@article{sparktts,
  title={Spark-tts: An efficient llm-based text-to-speech model with single-stream decoupled speech tokens},
  author={Wang, Xinsheng and Jiang, Mingqi and Ma, Ziyang and Zhang, Ziyu and Liu, Songxiang and Li, Linqin and Liang, Zheng and Zheng, Qixi and Wang, Rui and Feng, Xiaoqin and others},
  journal={arXiv preprint arXiv:2503.01710},
  year={2025}
}

@inproceedings{
streamingllm,
title={Efficient Streaming Language Models with Attention Sinks},
author={Guangxuan Xiao and Yuandong Tian and Beidi Chen and Song Han and Mike Lewis},
booktitle={The Twelfth International Conference on Learning Representations},
year={2024}
}

@inproceedings{shortgpt,
    title = "{S}hort{GPT}: Layers in Large Language Models are More Redundant Than You Expect",
    author = "Men, Xin  and
      Xu, Mingyu  and
      Zhang, Qingyu  and
      Yuan, Qianhao  and
      Wang, Bingning  and
      Lin, Hongyu  and
      Lu, Yaojie  and
      Han, Xianpei  and
      Chen, Weipeng",
    booktitle = "Findings of the Association for Computational Linguistics: ACL 2025",
    month = jul,
    year = "2025",
    pages = "20192--20204"
}

@inproceedings{
gromov2024unreasonable,
title={The Unreasonable Ineffectiveness of the Deeper Layers},
author={Andrey Gromov and Kushal Tirumala and Hassan Shapourian and Paolo Glorioso and Dan Roberts},
booktitle={The Thirteenth International Conference on Learning Representations},
year={2025}
}

@inproceedings{spade,
  title={Spade: Structured pruning and adaptive distillation for efficient llm-tts},
  author={Nguyen, Tan Dat and Kim, Jaehun and Kim, Ji-Hoon and Choi, Shukjae and Lim, Youshin and Chung, Joon Son},
  booktitle={2026 IEEE International Conference on Acoustics, Speech and Signal Processing (ICASSP)},
  pages={16697--16701},
  year={2026},
  organization={IEEE}
}

@inproceedings{linaspeech,
  title={Lina-Speech: Gated Linear Attention and Initial-State Tuning for Multi-Sample Prompting Text-To-Speech Synthesis},
  author={Lemerle, Th{\'e}odor and Guichoux, T{\'e}o and Roebel, Axel and Obin, Nicolas},
  booktitle ={Proceedings of the AAAI 2026 Workshop on Audio-Centric AI: Towards Real-World Multimodal Reasoning and Application Use Cases (Audio-AAAI)},
  series={Proceedings of Machine Learning Research},
  volume={312},
  pages={1--20},
  year={2026},
  publisher={PMLR}
}

@article{hinton2015distilling,
  title={Distilling the Knowledge in a Neural Network},
  author={Hinton, Geoffrey and Vinyals, Oriol and Dean, Jeff},
  journal={arXiv preprint arXiv:1503.02531},
  year={2015}
}

@inproceedings{leviathan2023fast,
  title={Fast inference from transformers via speculative decoding},
  author={Leviathan, Yaniv and Kalman, Matan and Matias, Yossi},
  booktitle={International Conference on Machine Learning},
  pages={19274--19286},
  year={2023},
  organization={PMLR}
}

@article{seedtts,
  title={Seed-TTS: A Family of High-Quality Versatile Speech Generation Models},
  author={Anastassiou, Philip and Chen, Jiawei and Chen, Jitong and Chen, Yuanzhe and Chen, Zhuo and Chen, Ziyi and Cong, Jian and Deng, Lelai and Ding, Chuang and Gao, Lu and others},
  journal={arXiv preprint arXiv:2406.02430},
  year={2024}
}

@inproceedings{libritts,
  title     = {{LibriTTS: A Corpus Derived from LibriSpeech for Text-to-Speech}},
  author    = {Heiga Zen and Viet Dang and Rob Clark and Yu Zhang and Ron J. Weiss and Ye Jia and Zhifeng Chen and Yonghui Wu},
  year      = {2019},
  booktitle = {{Interspeech 2019}},
  pages     = {1526--1530}
}

@article{deng2025indextts,
  title={Indextts: An industrial-level controllable and efficient zero-shot text-to-speech system},
  author={Deng, Wei and Zhou, Siyi and Shu, Jingchen and Wang, Jinchao and Wang, Lu},
  journal={arXiv preprint arXiv:2502.05512},
  year={2025}
}

@article{
defossez2022high,
title={High Fidelity Neural Audio Compression},
author={Alexandre D{\'e}fossez and Jade Copet and Gabriel Synnaeve and Yossi Adi},
journal={Transactions on Machine Learning Research},
issn={2835-8856},
year={2023}
}

@article{zeghidour2021soundstream,
  title={Soundstream: An end-to-end neural audio codec},
  author={Zeghidour, Neil and Luebs, Alejandro and Omran, Ahmed and Skoglund, Jan and Tagliasacchi, Marco},
  journal={IEEE/ACM Transactions on Audio, Speech, and Language Processing},
  volume={30},
  pages={495--507},
  year={2022},
  publisher={IEEE}
}

@InProceedings{park2024long,
  title = 	 {Long-Form Speech Generation with Spoken Language Models},
  author =       {Park, Se Jin and Salazar, Julian and Jansen, Aren and Kinoshita, Keisuke and Ro, Yong Man and Skerry-Ryan, Rj},
  booktitle = 	 {Proceedings of the 42nd International Conference on Machine Learning},
  pages = 	 {48245--48261},
  year = 	 {2025},
  volume = 	 {267}
}

@inproceedings{yang2024gated,
author = {Yang, Songlin and Wang, Bailin and Shen, Yikang and Panda, Rameswar and Kim, Yoon},
title = {Gated linear attention transformers with hardware-efficient training},
year = {2024},
publisher = {JMLR.org},
booktitle = {Proceedings of the 41st International Conference on Machine Learning},
articleno = {2333},
numpages = {23},
location = {Vienna, Austria},
series = {ICML'24}
}

@inproceedings{radford2023robust,
  title={Robust speech recognition via large-scale weak supervision},
  author={Radford, Alec and Kim, Jong Wook and Xu, Tao and Brockman, Greg and McLeavey, Christine and Sutskever, Ilya},
  booktitle={International conference on machine learning},
  pages={28492--28518},
  year={2023},
  organization={PMLR}
}

@article{chen2022wavlm,
  title={Wavlm: Large-scale self-supervised pre-training for full stack speech processing},
  author={Chen, Sanyuan and Wang, Chengyi and Chen, Zhengyang and Wu, Yu and Liu, Shujie and Chen, Zhuo and Li, Jinyu and Kanda, Naoyuki and Yoshioka, Takuya and Xiao, Xiong and others},
  journal={IEEE Journal of Selected Topics in Signal Processing},
  volume={16},
  number={6},
  pages={1505--1518},
  year={2022},
  publisher={IEEE}
}

@inproceedings{saeki2022utmos,
  title     = {{UTMOS: UTokyo-SaruLab System for VoiceMOS Challenge 2022}},
  author    = {Takaaki Saeki and Detai Xin and Wataru Nakata and Tomoki Koriyama and Shinnosuke Takamichi and Hiroshi Saruwatari},
  year      = {2022},
  booktitle = {{Interspeech 2022}},
  pages     = {4521--4525},
  doi       = {10.21437/Interspeech.2022-439},
  issn      = {2958-1796},
}

@article{pope2022efficiently,
  title={Efficiently scaling transformer inference},
  author={Pope, Reiner and Douglas, Sholto and Chowdhery, Aakanksha and Devlin, Jacob and Bradbury, James and Levskaya, Anselm and Heek, Jonathan and Xiao, Kefan and Agrawal, Shivani and Dean, Jeff},
  journal={Proceedings of machine learning and systems},
  volume={5},
  pages={606--624},
  year={2023}
}

@inproceedings{li2025fast,
  title={Fast and high-quality auto-regressive speech synthesis via speculative decoding},
  author={Li, Bohan and Wang, Hankun and Zhang, Situo and Guo, Yiwei and Yu, Kai},
  booktitle={2025 IEEE International Conference on Acoustics, Speech and Signal Processing (ICASSP)},
  year={2025},
  organization={IEEE}
}

@inproceedings{nguyen2025accelerating,
  title={Accelerating codec-based speech synthesis with multi-token prediction and speculative decoding},
  author={Nguyen, Tan Dat and Kim, Ji-Hoon and Choi, Jeongsoo and Choi, Shukjae and Park, Jinseok and Lee, Younglo and Chung, Joon Son},
  booktitle={2025 IEEE International Conference on Acoustics, Speech and Signal Processing (ICASSP)},
  pages={1--5},
  year={2025},
  organization={IEEE}
}

@inproceedings{mamba2,
  title={Transformers are {SSM}s: Generalized Models and Efficient Algorithms Through Structured State Space Duality},
  author={Dao, Tri and Gu, Albert},
  booktitle={International Conference on Machine Learning (ICML)},
  year={2024}
}

@inproceedings{DBLP:conf/icassp/JiangLFHM25,
  author       = {Xilin Jiang and
                  Yinghao Aaron Li and
                  Adrian Nicolas Florea and
                  Cong Han and
                  Nima Mesgarani},
  title        = {Speech Slytherin: Examining the Performance and Efficiency of Mamba
                  for Speech Separation, Recognition, and Synthesis},
  booktitle    = {2025 IEEE International Conference on Acoustics, Speech and Signal Processing (ICASSP)},
  pages        = {1--5},
  publisher    = {{IEEE}},
  year         = {2025}
}

@article{qwen2024qwen2,
  title={Qwen2.5 technical report},
  author={Yang, An and Yang, Baosong and Zhang, Beichen and Hui, Binyuan and Zheng, Bo and Yu, Bowen and Li, Chengpeng and Liu, Dayiheng and Huang, Fei and Wei, Haoran and others},
  journal={arXiv preprint arXiv:2412.15115},
  year={2024}
}

@inproceedings{funasr,
  title     = {{FunASR: A Fundamental End-to-End Speech Recognition Toolkit}},
  author    = {Zhifu Gao and Zerui Li and Jiaming Wang and Haoneng Luo and Xian Shi and Mengzhe Chen and Yabin Li and Lingyun Zuo and Zhihao Du and Shiliang Zhang},
  year      = {2023},
  booktitle = {{Interspeech 2023}},
  pages     = {1593--1597},
  doi       = {10.21437/Interspeech.2023-1428},
  issn      = {2958-1796},
}

@inproceedings{ainslie2023gqa,
    title = "{GQA}: Training Generalized Multi-Query Transformer Models from Multi-Head Checkpoints",
    author = "Ainslie, Joshua  and
      Lee-Thorp, James  and
      de Jong, Michiel  and
      Zemlyanskiy, Yury  and
      Lebron, Federico  and
      Sanghai, Sumit",
    booktitle = "Proceedings of the 2023 Conference on Empirical Methods in Natural Language Processing",
    year = "2023",
    pages = "4895--4901"
}

@InProceedings{ko2024distillm,
  title = 	 {{D}isti{LLM}: Towards Streamlined Distillation for Large Language Models},
  author =       {Ko, Jongwoo and Kim, Sungnyun and Chen, Tianyi and Yun, Se-Young},
  booktitle = 	 {Proceedings of the 41st International Conference on Machine Learning},
  pages = 	 {24872--24895},
  year = 	 {2024},
  volume = 	 {235},
  series = 	 {Proceedings of Machine Learning Research}
}

@article{nguyen2026mamtra,
  title={MamTra: A Hybrid Mamba-Transformer Backbone for Speech Synthesis},
  author={Nguyen, Tan Dat and Bae, Sangmin and Chung, Joon Son and Kim, Ji-Hoon},
  journal={arXiv preprint arXiv:2603.12342},
  year={2026}
}

@inproceedings{kwon2023efficient,
  title={Efficient Memory Management for Large Language Model Serving with PagedAttention},
  author={Woosuk Kwon and Zhuohan Li and Siyuan Zhuang and Ying Sheng and Lianmin Zheng and Cody Hao Yu and Joseph E. Gonzalez and Hao Zhang and Ion Stoica},
  booktitle={Proceedings of the ACM SIGOPS 29th Symposium on Operating Systems Principles},
  year={2023}
}

@inproceedings{wang2017tacotron,
  title={Tacotron: Towards End-to-End Speech Synthesis},
  author={Wang, Yuxuan and Skerry-Ryan, RJ and Stanton, Daisy and Wu, Yonghui and Weiss, Ron J and Jaitly, Navdeep and Yang, Zongheng and Xiao, Ying and Chen, Zhifeng and Bengio, Samy and others},
  booktitle={Proc. Interspeech 2017},
  pages={4006--4010},
  year={2017}
}

@inproceedings{kim2021conditional,
  title={Conditional variational autoencoder with adversarial learning for end-to-end text-to-speech},
  author={Kim, Jaehyeon and Kong, Jungil and Son, Juhee},
  booktitle={International conference on machine learning},
  pages={5530--5540},
  year={2021},
  organization={PMLR}
}

@inproceedings{santilli2023accelerating,
  title={Accelerating transformer inference for translation via parallel decoding},
  author={Santilli, Andrea and Severino, Silvio and Postolache, Emilian and Maiorca, Valentino and Mancusi, Michele and Marin, Riccardo and Rodol{\`a}, Emanuele},
  booktitle={Proceedings of the 61st Annual Meeting of the Association for Computational Linguistics (Volume 1: Long Papers)},
  pages={12336--12355},
  year={2023}
}

@inproceedings{kou2024cllms,
  title={Cllms: Consistency large language models},
  author={Kou, Siqi and Hu, Lanxiang and He, Zhezhi and Deng, Zhijie and Zhang, Hao},
  booktitle={Forty-first International Conference on Machine Learning},
  year={2024}
}

@inproceedings{kim2024infinipot,
  title={Infinipot: Infinite context processing on memory-constrained llms},
  author={Kim, Minsoo and Shim, Kyuhong and Choi, Jungwook and Chang, Simyung},
  booktitle={Proceedings of the 2024 Conference on Empirical Methods in Natural Language Processing},
  pages={16046--16060},
  year={2024}
}

@inproceedings{liu2025reattention,
  title={{ReAttention}: Training-free infinite context with finite attention scope},
  author={Liu, Xiaoran and Li, Ruixiao and Liu, Zhigeng and Guo, Qipeng and Song, Yuerong and Lv, Kai and Yan, Hang and Li, Linlin and Liu, Qun and Qiu, Xipeng},
  booktitle={International Conference on Learning Representations},
  year={2025}
}

@article{kang2026unlocking,
  title={Unlocking Fine-Grained and Within-Utterance Speaking Style Control in Prompt-Based Text-to-Speech Models},
  author={Kang, Jaehoon and Lee, Yejin and Park, Yoonji and Shim, Kyuhong},
  journal={arXiv preprint arXiv:2605.27376},
  year={2026}
}

@inproceedings{yoon2023pruning,
  title     = {{Pruning Self-Attention for Zero-Shot Multi-Speaker Text-to-Speech}},
  author    = {Hyungchan Yoon and Changhwan Kim and Eunwoo Song and Hyun-Wook Yoon and Hong-Goo Kang},
  year      = {2023},
  booktitle = {{Interspeech 2023}},
  pages     = {4299--4303},
  doi       = {10.21437/Interspeech.2023-1301},
  issn      = {2958-1796},
}

@article{inaguma2021alignment,
  title={Alignment knowledge distillation for online streaming attention-based speech recognition},
  author={Inaguma, Hirofumi and Kawahara, Tatsuya},
  journal={IEEE/ACM Transactions on Audio, Speech, and Language Processing},
  volume={31},
  pages={1371--1385},
  year={2023},
  publisher={IEEE}
}

@inproceedings{yoon2025heuristic,
  title={Heuristic-free knowledge distillation for streaming ASR via multi-modal training},
  author={Yoon, Ji Won},
  booktitle={Proceedings of the AAAI Conference on Artificial Intelligence},
  volume={39},
  number={24},
  pages={25733--25741},
  year={2025}
}

@string{aaai =  "Proc. AAAI"}

@string{icassp="Proc. ICASSP"}

@string{icml =  "Proc. ICML"}

@string{is =  "Proc. Interspeech"}

\end{document}